\documentclass[runningheads]{llncs}
\usepackage[T1]{fontenc}
\usepackage{amsmath,amsfonts,bm}
\usepackage{hyperref}
\usepackage{url}
\usepackage{cleveref}
\usepackage{graphicx}
\usepackage{bm}
\usepackage{booktabs}

\DeclareMathOperator*{\argmaxB}{argmax}

\DeclareMathOperator*{\topfive}{top-5}
\newcommand{\mname}{AutoXPCR}

\begin{document}
\title{AutoXPCR: Automated Multi-Objective Model Selection for Time Series Forecasting}
\titlerunning{Automated Multi-Objective Model Selection for Time Series Forecasting}
%
\author{Raphael Fischer\orcidID{0000-0002-1808-5773} \and
Amal Saadallah\orcidID{0000-0003-2976-7574} }
\authorrunning{Fischer and Saadallah} 
%
\institute{Lamarr Institute for Machine Learning and Artificial Intelligence\\TU Dortmund University, 44227 Dortmund, Germany\\
\email{firstname.lastname@tu-dortmund.de}\\\url{https://lamarr.cs.tu-dortmund.de/}} 
\maketitle              
\begin{abstract}
Automated machine learning (AutoML) streamlines the creation of ML models. While most methods select the ``best'' model based on predictive quality, it's crucial to acknowledge other aspects, such as interpretability and resource consumption. This holds particular importance in the context of deep neural networks (DNNs), as these models are often perceived as computationally intensive black boxes. In the challenging domain of time series forecasting, DNNs achieve stunning results, but specialized approaches for automatically selecting models are scarce. In this paper, we propose \mname~-- a novel method for automated and explainable multi-objective model selection. Our approach leverages meta-learning to estimate any model's performance along PCR criteria, which encompass (P)redictive error, (C)omplexity, and (R)esource demand. Explainability is addressed on multiple levels, as our interactive framework can prioritize less complex models and provide by-product explanations of recommendations. We demonstrate practical feasibility by deploying \mname~on over 1000 configurations across 114 data sets from various domains. Our method clearly outperforms other model selection approaches -- on average, it only requires 20\% of computation costs for recommending models with 90\% of the best-possible quality.

\end{abstract}

\section{Introduction}

\begin{figure}
    \centering
    \includegraphics[width=.8\columnwidth]{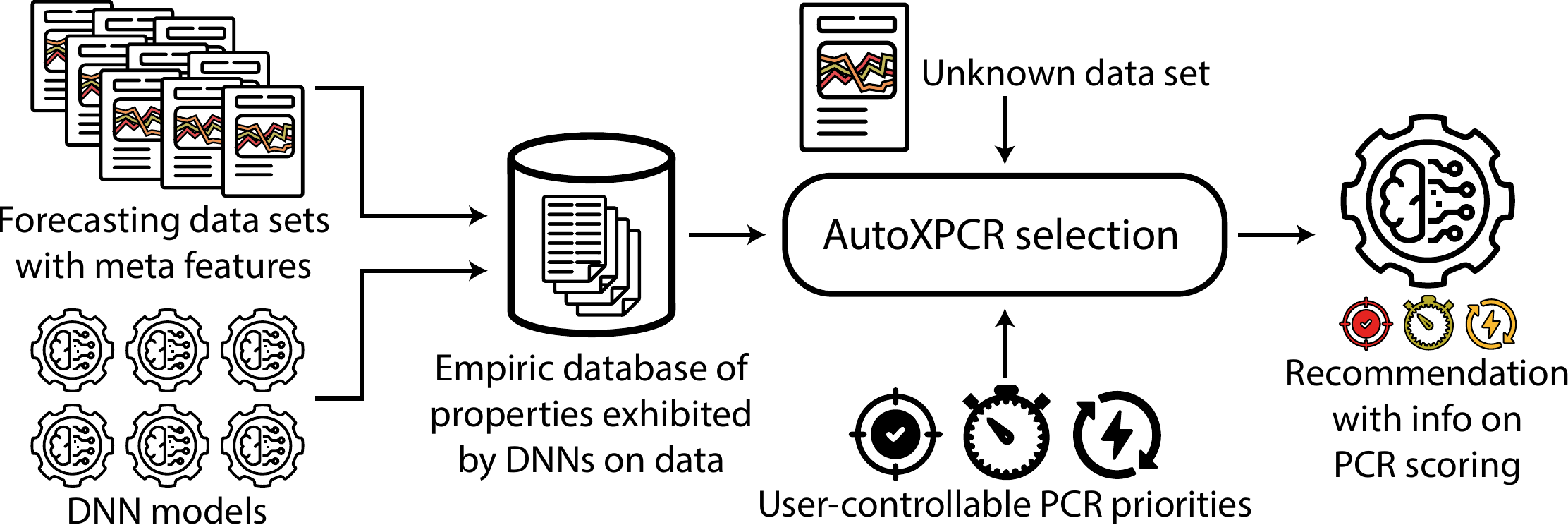}
    \caption{Framework for \mname, which leverages meta-learning to estimate DNN performance on given data in terms of prediction error, complexity, and resource demand. The recommendation has the best-estimated trade-off among all properties.}
    \label{fig:famework}
\end{figure}

The rapidly evolving field of machine learning (ML) brought forth a broad arsenal of methods that exhibit properties related to predictive performance, model complexity, and computational resource demand.
Deployment requires evaluating the pros and cons of applicable methods, and as there is ``no free lunch'', these trade-offs vary across data sets \cite{Assessing_Energy_Efficiency_of_ML}.
Competing properties can be found across all learning tasks, including time series forecasting, which is challenging due to the complex and time-evolving nature of data \cite{saadallah/etal/2019}.
Deep neural networks (DNNs) have proven to be highly potent for this task \cite{saadallah/etal/2022a,deeparsalinas2020}. However, their complex structure makes them computationally expensive \cite{Strubell/etal/2020a} and non-transparent -- resulting in them being referred to as black boxes \cite{molnar2020interpretable}.
As a result, forecasting with DNNs necessitates explicitly investigating the aforementioned trade-offs.

The relevance of individual model properties is subject to the application and user requirements at hand.
For instance, consider edge devices \cite{buschjager2020site}, where computational resource demand is tightly constrained \cite{morik2022machine}.
Thus, selecting the ``best'' model can be formulated as a \emph{multi-objective optimization problem}, where objectives represent the properties as prioritized in the given ML use case.
Even with limiting the model search space to DNNs, the options for forecasting are vast \cite{gluontsjmlr}.
Practically, this renders exhaustive search (i.e., identifying the optimal model by testing every option) computationally redundant and, quite possibly, unfeasible.
Benchmarks such as the Monash repository \cite{godahewa2021monash} come to aid by listing empiric state-of-the-art (SOTA) results.
However, they usually overly focus on predictive performance and fail to report resource consumption. 
Automated ML (AutoML) pipelines \cite{jin2019auto,abdallah/etal/22} suffer from the same phenomenon and, moreover, are usually not specialized for time series \cite{alsharef2022review}.
In addition, most AutoML methods work in a non-transparent fashion, which is problematic considering the acknowledged significance of explainability \cite{molnar2020interpretable,rojat2021explainable}.

To overcome these challenges, our work proposes the framework of \mname, which is schematically displayed in Figure \ref{fig:famework}.
It is a novel approach for selecting DNN models to forecast time series in an automated, explainable, and resource-aware fashion.
We leverage multi-objective meta-learning, which considers time series' characteristics as features and selects models based on their suitability in terms of \emph{(P)rediction error}, \emph{(C)omplexity}, and \emph{(R)esource consumption}.
As probing the underlying optimization problem's Pareto front is computationally demanding, \mname~offers a more efficient solution by first estimating any model's fit along the objectives from prior knowledge.
Our method is explainable on three levels:
(1) it accompanies every recommendation with scoring along all PCR criteria,
(2) it only uses fully interpretable meta-learners that provide by-product explanations \cite{rudin2019stop}, and
(3), it provides means for interactive control of the process.
In addition to making our method more explainable, this enables users to favor more resource-aware or interpretable DNNs, allowing them to align the impact of each PCR criterion on the overall assessment with their priorities.
To validate our method, we provide an extensive experimental evaluation which showcases the effectiveness of \mname.
We applied 11 SOTA DNNs to over 100 data sets and perform meta-learning on the results from these configurations (1254 in total) -- the complete implementation and logs are publicly available at \url{github.com/raphischer/xpcr}.
Our method beats competing approaches and achieves 90\% of predictive performance at only 20\% of the computation cost.
Our work contributes to research on time series forecasting, automated model selection, meta-learning, explainability, and resource-aware ML.

\section{Related Work}
\label{sec:rw}

Before explaining the intricacies of our approach, we discuss the literary background of the research our work builds on, starting with deep learning in the context of forecasting \cite{saadallah/etal/2022a}.
The domain is currently dominated by recurrent (RNNs), particularly long short-term memory (LSTM) neural networks \cite{deeparsalinas2020}.
Convolutional networks (CNNs) also score well in forecasting and tend to be more efficient than RNNs \cite{borovykh2017conditional}.
The GluonTS toolkit \cite{gluontsjmlr} offers a broad range of DNN models for usage on time series.
DeepAR employs an RNN with LSTM or Gated Recurrent Units \cite{deeparsalinas2020}, while DeepState learns RNN weights jointly across all time series via a Kalman filter \cite{deepstate2018rangapuram}.
MQ-RNN and MQ-CNN leverage RNN and dilated causal CNN encoders, respectively, coupled with a quantile decoder, allowing sequence-to-sequence prediction \cite{wen2017multi}.
DeepFactor also estimates weights across series and combines local and global aspects \cite{deepfactor2019wang}.
GluonTS also offers transformer architectures \cite{tempfus2021lim,oord2016wavenet}, which are popular in language processing \cite{Strubell/etal/2020a}.

Utilizing ML for any task requires selecting the most promising model from the range of candidates, for which multiple approaches exist.
Firstly, one can try to estimate each candidate's expected error, e.g., via Gaussian \cite{birge2001gaussian} or Bayesian estimation methods.
These methods are however impractical for forecasting since they require approximating continuous composite densities for the error between target and estimated values.
Each candidate's error can also be estimated from additional models trained on prior empirical evaluations, thus implementing the idea of meta-learning \cite{wolpert1992stacked}.
Naturally, these methods are limited to the empirically investigated data, methods, and properties.
For forecasting, Godahewa et al. provided a good benchmark \cite{godahewa2021monash}, and meta-learning has been successfully adapted \cite{abdallah/etal/22,saadallah/etal/2019,saadallah/etal/2022a}.
AutoML can be understood as meta-learning without explicitly formulating meta-features
While being popular in deep learning \cite{jin2019auto}, there exist few approaches \cite{lyu2022one} that have specifically considered the sequential dependencies, variable-length inputs, and dynamic patterns faced in forecasting.
In addition to the aforementioned drawbacks, established model selection approaches suffer from exclusively focusing on predictive performance and neglecting properties relating to trustworthiness and resource consumption.

Many works have highlighted the importance of trustworthiness \cite{Brundage/etal/2020a}, which was eventually also manifested in the EU AI act \cite{eu-ethics}.
Research on explainability \cite{molnar2020interpretable} aims at making complex models like DNNs more interpretable \cite{rudin2019stop}.
It is also of particular importance to report on sustainability and resource-awareness \cite{DBLP:journals/corr/abs-2104-10350}, as the SOTA in ML domains like computer vision \cite{Schwartz/etal/2020b} and language processing \cite{DBLP:journals/corr/abs-2104-10350} was shown to significantly impact our environment \cite{Strubell/etal/2020a}.
As a result, the ML community is expected to establish practices for reducing carbon emissions \cite{Patterson2022,lacoste2019quantifying} and explicitly report on resource efficiency \cite{Assessing_Energy_Efficiency_of_ML}.
With ML becoming popular for business use \cite{fischer_prioritization_2023}, works have also motivated to specifically bridge the communication gap towards non-experts \cite{yeswecare,Assessing_Energy_Efficiency_of_ML}.
On the contrary, many SOTA works (including model selection) still overly focus on predictive performance \cite{Schwartz/etal/2020b} and fail to discuss broader, societal implications.
Our work aims to increase the trustworthiness and resource efficiency of forecasting with ML by explicitly embedding these aspects into our method, which we will now discuss in detail.

\section{Explainable Multi-Objective Model Selection}
Any ML experiment is characterized by an underlying configuration and environment \cite{Assessing_Energy_Efficiency_of_ML}.
The former specifies the model, data set, and learning task, while the latter represents the software and hardware platform for practical execution.
While we specifically formulate our methodology for the task of \emph{learning and evaluating a DNN on forecasting data}, it can be easily generalized to other ML domains.
Figure \ref{fig:famework} functions as a useful guide through our methodology.

\subsection{Problem Statement}
\label{subs:meth:problem}
Let $Y = \{y_1, y_2, \cdots, y_t\}$ be a time series, i.e., a sequence of $t$ recorded values $y_i$. 
Usually, such data is recorded in a wider context $\mathbf{Y}$, which encloses simultaneous time variables.
For example, consider weather data, where a single series $Y \in \mathbf{Y}$ might represent temperature, wind speed, or humidity recordings.
Univariate forecasting predicts next values based on recently observed data of a single series.
We understand automated selection of the best model $m$ from a pool of DNNs $M$ for forecasting these series as a multi-objective optimization problem \cite{YANG2014197}:
\begin{equation}\label{eq:optimization}
\argmaxB_{m \in M} ( F(\mathbf{Y},m) ) = \argmaxB_{m \in M} ( \sum_{i=1}^k w_i f_i(\mathbf{Y},m) )\textrm{ with } \forall i, w_i\geq0  \textrm{ and }  \sum_{i=1}^k w_i=1
\end{equation}
The \emph{compound score} $F(\mathbf{Y},m)$ is a weighted sum of functions $f_{i}$, which describe properties that $m$ exhibits when forecasting $\mathbf{Y}$.
As properties can be partitioned into groups describing (P)rediction error, (C)omplexity, or (R)esource consumption, we name the $f_{i}$ \emph{PCR functions}. 
For example, one could assess the error on test data (P), model size (C), or power draw during inference (R).
Note that C and R properties are directly linked to aspects of explainability and resource-awareness, which as earlier motivated need to be explicitly considered.
Following Equation \ref{eq:optimization}, the individual PCR functions should be defined such that maximization leads to improved model behavior.
The weights $w_{i}$ control each property's impact on the compound score, which allows to explicitly favor resource-friendly or explainable models.
Note that solving Equation \ref{eq:optimization} efficiently is non-trivial:
Firstly, the functions $f_i$ can behave in a contradictory manner, which complicates simultaneous optimization -- to give an example, it is hard to maintain low prediction errors when making models less complex.
As a result, there might not even be a single solution to our problem but rather a range of Pareto-optimal choices.
Equation \ref{eq:optimization} can be naively solved by performing an \emph{exhaustive search}, i.e. determining the PCR properties of all DNN options $m \in M$ by training and evaluating them on $\mathbf{Y}$.
While guaranteed to provide optimal solutions, this also requires dramatic computational expenses.
Using classic optimization to navigate the Pareto front of model choices \cite{YANG2014197} is also unfeasible, as the PCR functions $f_i$ are non-differentiable -- values can be solely obtained from deploying models $m$ on $\mathbf{Y}$.
Actually, even testing just a single model to assess its properties unveils a major problem, which we first need to tackle.

\subsection{Comparability of Properties}
\label{ssec:meth:comparability}
The meaning of properties and numeric values will be vastly different -- we could for example register hundreds of milliseconds for running time, dozens of kilowatt-hours for power draw, or millions of parameters.
This problem gets more evident when considering different environments for running models (e.g., CPU or GPU implementations), as this choice will dramatically change the value magnitudes.
We address this issue by assessing values of $f_{i}$ on a relative \emph{index scale}, based on the real measurements $\mu_i$ obtained from forecasting $\mathbf{Y}$ with $m$ \cite{Assessing_Energy_Efficiency_of_ML}.
Whereas the original work calculates index values based on reference models, we instead resort to putting values in relation to the empirically best-performing model $m^\ast_i$ on the $i$-th property:
\begin{equation}\label{eq:index}
    f_i(\mathbf{Y},m)=\left(\frac{\mu^*_i(\mathbf{Y})}{\mu_i(\mathbf{Y},m)}\right)=\left(\frac{ \min_{m^\ast_i\in M}( \mu_i(\mathbf{Y},m^\ast_i))}{\mu_i(\mathbf{Y},m)}\right),
\end{equation}
The now calculable PCR function values $f_{i}$ and compound score $F$ are bounded by the interval $(0, 1]$, making them more easily comparable.
They describe the behavior in a given environment relatively; the higher the value, the closer it is to the best empiric result which receives $f_i(\mathbf{Y}, m^\ast_i) = 1$.
Note that using the best value per property as a reference is advantageous to global reference models \cite{Assessing_Energy_Efficiency_of_ML}, since it neatly solves the problem of choosing these references.

\subsection{Model Selection using Meta-learning}
\label{ssec:meth:modelselection}
To solve Equation \ref{eq:optimization} efficiently, we propose to adapt meta-learning.
It allows to estimate the values of $f_i$ given the characteristics of input data $\mathbf{Y}$ and model $m$ instead of applying it practically.
To put it clearly, we understand the meta-task as training regression models $\hat{f}_{i}$, which estimate PCR function values $\hat{f}_{i}(X_{\mathbf{Y}}, X_m)$ from meta-features for time series $X_{\mathbf{Y}}$ and DNNs $X_m$.
The former $X_{\mathbf{Y}} \in \mathcal{F}_{\mathbf{Y}}, \mathcal{F}_{\mathbf{Y}} \subset \mathbb{R}^n$ describe the temporal data and thus might encode information like statistical numbers, seasonality, or stationarity.
The model meta-features $X_m \in \mathcal{F}_{m}, \mathcal{F}_{m} \subset \mathbb{R}^{n^\prime}$ encode general information about any DNN, such as the number of network layers and their type.
Details about the meta-features used in our experiments are provided in Section \ref{sec:exp}.
Accordingly, the meta-learners $\hat{f}_{i} \in \mathcal{F}$ map from the feature spaces onto real values, i.e., $\mathcal{F}: \mathcal{F}_{\mathbf{Y}} \times \mathcal{F}_{m} \rightarrow \mathbb{R}$.

Training the regressors requires to collect a \emph{property database} $D$ of meta-features and real function values, i.e., $D = \{(X_{\mathbf{Y}}, X_m;f_i(\mathbf{Y},m))\}$.
With the aforementioned ``no free lunch'' theorem in mind, cross-validation across the database allows to identify the best regression method for each individual property, i.e., the one with the lowest estimation error (more meta-learner quality indicators are given in Table \ref{tab:metalearn_errors}):
\begin{equation}\label{eq:estimation_error}
    \min_{\hat{f}_{i}} \sum_{(X_{\mathbf{Y}}, X_m;f_i(\mathbf{Y},m)) \in D} |f_i(\mathbf{Y},m) - \hat{f}_{i}(X_{\mathbf{Y}}, X_m)|
\end{equation}

Given the property weights and meta-features of any data set, we can now automatically recommend models by estimating solutions for Equation \ref{eq:optimization}.
For this, we replace $f_{i}$ with the meta-learner output $\hat{f}_{i}$ and run queries for the meta-features of all model choices (instead of applying them to $\mathbf{Y}$).
The resulting estimated compound score $\hat{F}(x)$ indicates each model's performance under consideration of all PCR aspects, hence the name \mname.
As an alternative approach, one could also \emph{directly} meta-learn the compound scores $F$.
However, this would not allow for estimating or weighting the individual PCR properties, which makes the method less transparent and interactive.
We later show empiric evidence that this approach - as expected - does not perform better than a \emph{compositional}, PCR-aware recommendation, where outputs of individually trained meta-regressors are aggregated via Equation \ref{eq:optimization}.

\subsection{Explainability Aspects}
We argue \mname to be explainable, as it offers insights on thee distinct levels.
Firstly, for any query, our method provides by-product explanations in the form of estimates for all property functions.
They inform interested users about the recommendation's estimated PCR trade-offs, explaining to what extent this model is expected to exhibit each property.
Thanks to the relative index scales (recall Equation \ref{ssec:meth:comparability}), this information is highly comprehensible in itself -- as an example, a score of $0.4$ implies that this model achieves $40\%$ of the best empirical result observed in $D$.
Secondly, decisions of \mname provide additional explanations, thanks for our restriction of only considering fully interpretable ML methods \cite{rudin2019stop} as meta-learners.
As such, the PCR function regressors inform on feature importance, which enables users to understand the link between model recommendations and certain data or model characteristics.
Performing a sensitivity analysis would allow us to go even deeper and explore how changes to the meta-features would impact the recommendation.
Lastly, interactions have been shown to potentially improve trust and explainability \cite{beckhsok}, which also applies to our framework.
By interactively weighting the individual properties, the selection process becomes controllable and more transparent.
As an additional benefit, it enables users to specifically prioritize less complex, or in other words, more interpretable models, and be resource-aware when investigating model choices.
Lastly, we followed the call for making methods results more comprehensible to non-experts via discrete ratings and informative labels \cite{yeswecare,Assessing_Energy_Efficiency_of_ML} - this is readily supported by our framework implementation.

\section{Experiments}
\label{sec:exp}

\setlength{\tabcolsep}{2.4pt}

\begin{table}[t]
    \begin{minipage}[t]{.598\linewidth}
        \caption{Forecasting properties, associated group, and corresponding impact on compound score}
        \label{tab:properties}
        \renewcommand{\arraystretch}{0.5}
        \begin{tabular}{lcc}
            Property & Group & Weight \\
            \midrule
            Test MASE & Performance & 0.111 \\
            Test RMSE & Performance & 0.111 \\
            Test MAPE & Performance & 0.111 \\
            Number of Parameters & Complexity & 0.167 \\
            Model Size on Disc & Complexity & 0.167 \\
            Training Power Draw & Resources & 0.083 \\
            Training Time & Resources & 0.083 \\
            Power Draw per Inference & Resources & 0.083 \\
            Running Time per Inference & Resources & 0.083 \\
        \end{tabular}
    \end{minipage}%
    \hfill
    \begin{minipage}[t]{.36\linewidth}
        \caption{Error measures for assessing meta-learner quality}
        \label{tab:metalearn_errors}
        \renewcommand{\arraystretch}{1.5}
        \begin{tabular}{lc}
            (a) & $\epsilon = |f-\hat{f}|$ \\
            (b) & $\epsilon \stackrel{!}{<} 0.1$ \\
            (c) & $\argmaxB_{m}f \stackrel{!}{=} \argmaxB_{m}\hat{f}$ \\
            (d) & $\argmaxB_{m}f \in \{\topfive_{m}\hat{f}\}$ \\
            (e) & $\{\topfive_{m}f\} \cap \{\topfive_{m}\hat{f}\}$ \\
        \end{tabular}
    \end{minipage}
\end{table}

We now investigate the practicability of our method across 114 data sets, which consist of $5+1$ (five subsampled and the full) versions of 19 original Monash data sets.
11 popular DNNs were applied to each variant, namely Feed-Forward (FFO), DeepAR (DAR), N-BEATS (NBE), WaveNet (WVN), DeepState (DST), DeepFactor (DFA), Deep Renewal Processes (DRP), GPForecaster (GPF), MQ-CNN \& MQ-RNN (MQC \& MQC) and Temporal Fusion Transformer (TFT), as introduced in Section \ref{sec:rw}.
Our meta-features constitute information on seasonality and forecast horizon (as given by Monash \cite{godahewa2021monash}), averaged statistics across series, as well as one-hot encoded model choice, totaling in a database shape of $(1254, 21)$.
For meta-learning, we split the data via five-fold grouped cross-validation, i.e., all variants of an original data set are either used for training or validation.
Only simple, interpretable models \cite{rudin2019stop} like linear regressors, support vector regressors, and decision trees were used as meta-learners.
Table \ref{tab:properties} lists the PCR properties for Equation \ref{eq:optimization}, which assess model behavior.
Besides root mean squared error (RMSE), we investigate the mean absolute scaled (MASE) and percentage (MAPE) error as specialized metrics for forecasting \cite{hyndman2006another}.
The chosen weights $w_i$ mitigate correlations (e.g., between errors) but maintain a sound trade-off -- in sum, each PCR group is equally weighted.
As competitors, we tested an exhaustive search, AutoForecast \cite{abdallah/etal/22}, and AutoKeras \cite{jin2019auto} -- we did not find public code for other methods \cite{lyu2022one}.
Our code uses GluonTS \cite{gluontsjmlr} for deep learning, CodeCarbon \cite{codecarbon} for profiling, and Scikit-learn \cite{scikit-learn} for meta-learning.
Experiments were performed on a single PC (Xeon W-2155 CPU) and took about two weeks, resulting in estimated carbon emissions of 24 CO$_2$e \cite{lacoste2019quantifying}.
Note that the computational effort stems mostly from deploying the $1254$ DNNs -- running \mname~(i.e., training and validating the meta-regressors) only takes a few seconds.
Complete code and results are available at \url{github.com/raphischer/xpcr}, including a tool for interactively exploring our findings.

\begin{table*}[t]
    \renewcommand{\arraystretch}{0.5}
    \caption{Compound score F(x) (higher is better) and scaled error [MASE] of obtained models from different search approaches, along with power draw [kWh] of the method.}
    \label{tab:modelselect_comparison}
    \begin{center}
        \begin{tabular}{l|ccc|ccc||ccc||cc}
            \toprule 
            Data set & \multicolumn{3}{c}{AutoXPCR} & \multicolumn{3}{c}{AutoForecast} & \multicolumn{3}{c}{Exhaustive} & \multicolumn{2}{c}{AutoKeras} \\
              & F(x) & MASE & kWh & F(x) & MASE & kWh & F(x) & MASE & kWh & MASE & kWh \\
            \midrule
            Aust..and & \textbf{0.70} & \textbf{1.24} & \textbf{0.98} & 0.21 & 1.69 & 29.7 & \textbf{0.70} & \textbf{1.24} & 914 & 13.0 & 69.9 \\
            Car Parts & 0.66 & 0.50 & \textbf{0.70} & 0.38 & \textbf{0.47} & 1.55 & \textbf{0.67} & 0.75 & 56.3 & 0.97 & 8.47 \\
            CIF 2016 & \textbf{0.78} & 1.08 & \textbf{0.07} & 0.27 & \textbf{1.03} & 0.33 & \textbf{0.78} & 1.08 & 9.86 & 16.2 & 2.46 \\
            Dominick & \textbf{0.74} & \textbf{1.55} & \textbf{13.3} & 0.49 & 1.77 & 28.9 & \textbf{0.74} & \textbf{1.55} & 1000 & 2.53 & 245 \\
            Elec..kly & \textbf{0.76} & \textbf{1.65} & \textbf{0.42} & 0.39 & 2.79 & 0.97 & \textbf{0.76} & \textbf{1.65} & 52.5 & 19.8 & 12.2 \\
            FRED-MD & \textbf{0.73} & \textbf{0.71} & \textbf{0.15} & 0.26 & 1.92 & 0.57 & \textbf{0.73} & \textbf{0.71} & 12.3 & 991 & 6.32 \\
            Hospital & \textbf{0.72} & 0.84 & \textbf{0.96} & 0.45 & \textbf{0.76} & 3.92 & \textbf{0.72} & 0.84 & 52.2 & 75.5 & 20.0 \\
            M1 M..hly & 0.62 & \textbf{1.40} & \textbf{0.76} & 0.39 & 1.50 & 4.09 & \textbf{0.70} & 1.67 & 55.3 & 1172 & 21.6 \\
            M1 Q..rly & 0.60 & 3.15 & \textbf{0.30} & 0.37 & \textbf{1.78} & 0.85 & \textbf{0.68} & 1.99 & 16.5 & 335 & 2.23 \\
            M3 M..hly & \textbf{0.61} & \textbf{1.12} & \textbf{2.32} & 0.38 & 1.16 & 11.1 & \textbf{0.61} & \textbf{1.12} & 273 & 2.57 & 80.6 \\
            M3 Q..rly & \textbf{0.58} & \textbf{1.41} & \textbf{1.90} & 0.35 & 2.08 & 3.20 & \textbf{0.58} & \textbf{1.41} & 86.2 & 2.67 & 7.81 \\
            M4 Hourly & \textbf{0.66} & 5.09 & \textbf{0.48} & 0.22 & \textbf{3.14} & 7.16 & \textbf{0.66} & 5.09 & 414 & 7.7e4 & 213 \\
            M4 Weekly & 0.69 & \textbf{2.79} & \textbf{1.31} & 0.39 & 4.13 & 4.52 & \textbf{0.70} & 3.10 & 71.0 & 25.7 & 24.6 \\
            NN5 Daily & \textbf{0.66} & \textbf{0.65} & \textbf{0.23} & 0.42 & 0.68 & 0.80 & \textbf{0.66} & \textbf{0.65} & 42.7 & 1.21 & 29.7 \\
            NN5 ..kly & \textbf{0.74} & 0.95 & \textbf{0.19} & 0.38 & \textbf{0.88} & 1.56 & \textbf{0.74} & 0.95 & 38.9 & 1.58 & 2.16 \\
            Sola..kly & \textbf{0.70} & 1.47 & \textbf{0.10} & 0.44 & \textbf{0.83} & 0.48 & \textbf{0.70} & 1.47 & 23.9 & 1.00 & 1.24 \\
            Tour..hly & 0.59 & 1.70 & \textbf{0.48} & 0.34 & \textbf{1.44} & 3.30 & \textbf{0.64} & 2.79 & 58.5 & 6.94 & 41.3 \\
            Tour..rly & \textbf{0.77} & 1.96 & \textbf{0.27} & 0.48 & \textbf{1.67} & 0.66 & \textbf{0.77} & 1.96 & 24.2 & 12.9 & 11.9 \\
            Traf..kly & \textbf{0.77} & 1.69 & \textbf{0.66} & 0.34 & \textbf{1.53} & 5.72 & \textbf{0.77} & 1.69 & 107 & 5.06 & 9.74 \\
            \bottomrule
        \end{tabular}
    \end{center}
\end{table*}

Let us first investigate how \mname~performs compared to other approaches.
Table \ref{tab:modelselect_comparison} lists the compound score and MASE of each recommended model, as well as the power draw for running the model selection (recall Section \ref{ssec:meth:comparability} -- higher compound scores indicate better model performance).
As expected, the exhaustive search provides the best-trading model but requires immense computational effort.
AutoForecast \cite{abdallah/etal/22}, which can be understood as running \mname~with zero weights for all properties except MASE, on average recommends more accurate models.
However, the lower compound scores indicate that obtained models are inefficient, which can also be seen by the power draw of testing the candidate.
AutoKeras \cite{jin2019auto} fails at retrieving models of comparably good quality while demanding even higher amounts of power.
As AutoKeras is not specifically designed for forecasting, we did not include these obtained DNNs in our meta-learning database or assess their compound score.
In contrast, testing the top recommendation of \mname~has a much lower power draw and provides reasonably accurate models across all data sets.
In practice, our method seems to be highly energy efficient and clearly outperforms all competitors -- in most cases, it correctly predicts the best-trading model.

\begin{figure*}[t]
    \centering
    \includegraphics[width=\textwidth]{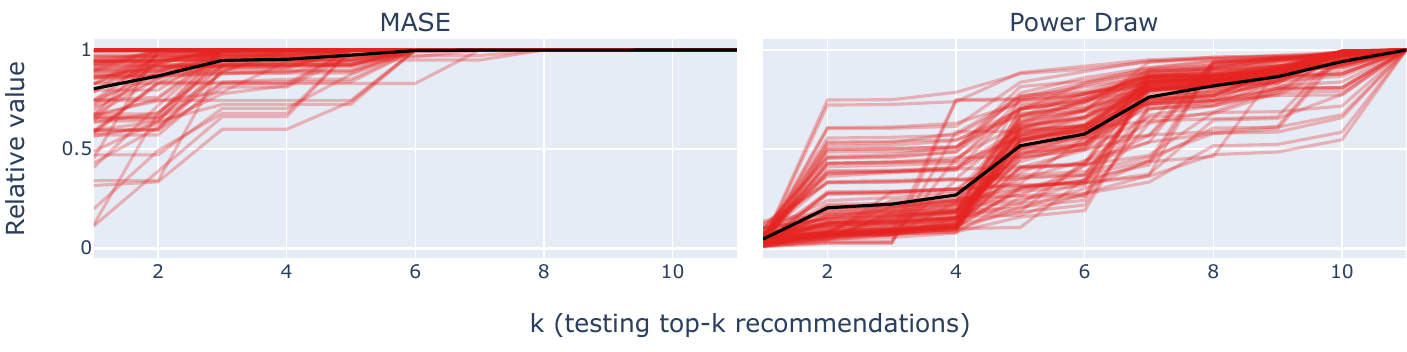}
    \caption{MASE and power draw of testing the top-$k$ recommended models for all data sets (red) and averaged (black), in relation to exhaustively testing all models. Trying two models results in an average relative MASE of 90\% at less than 20\% of the cost. }
    \label{fig:convergence}
\end{figure*}

\begin{figure}[t]
    \centering
    \includegraphics[width=.325\columnwidth]
    {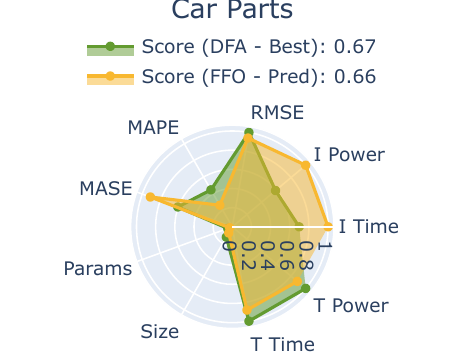}
    \includegraphics[width=.325\columnwidth]
    {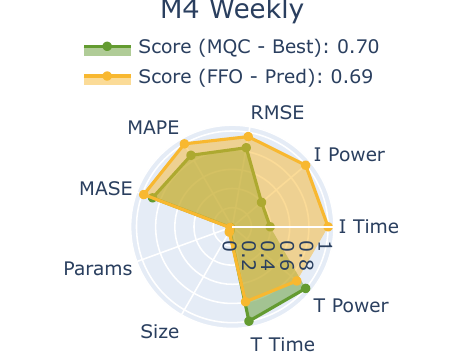}
    \includegraphics[width=.325\columnwidth]
    {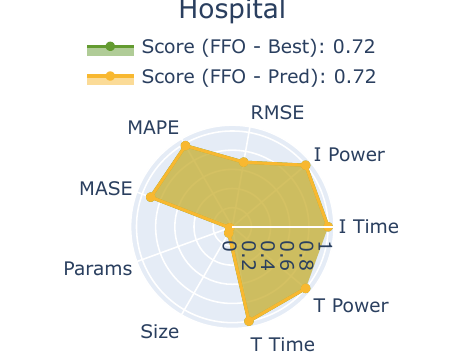}
    \caption{Properties of optimal and best-recommended models. On Car Parts and M4 Weekly, individual properties diverge, but the compound scores are very close.}
    \label{fig:optimal_recomm_comparison}
\end{figure}

By testing the top-$k$ recommended models, we can evaluate how \mname~converges towards optimal results.
This is visualized for all 114 data sets (red) in Figure \ref{fig:convergence}, with MASE and power draw being compared relatively to exhaustively testing all DNNs.
On average (black), ~90\% of the best possible MASE can already be achieved by just testing the top two recommendations, which requires less than ~20\% of the resources.
In over 95\% of cases, the optimal model will be under the top-5 recommendations.
We visualize the multi-objective trade-offs of actual best and recommended DNNs via star plots in Figure \ref{fig:optimal_recomm_comparison}.
Here, to obtain diverging star shapes, we purposefully selected two data sets where \mname~fails to recommend the best model.
It shows how both models behave quite differently but still score well on Equation \ref{eq:optimization}.

\begin{figure}[t]
    \centering
    \includegraphics[width=.495\columnwidth]
    {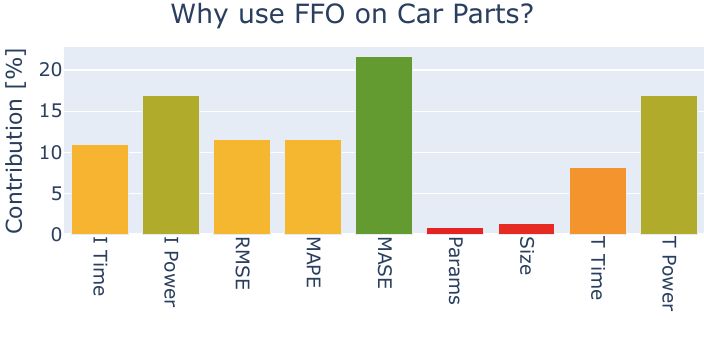}
    \includegraphics[width=.495\columnwidth]
    {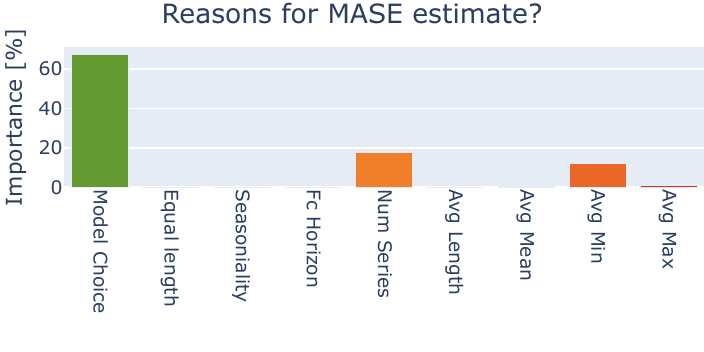}
    \caption{Explanations for model recommendation based on normalized contribution to the compound estimate (left) and feature importance for estimating MASE (right).}
    \label{fig:explanations}
\end{figure}

\begin{figure*}
    \centering
    \includegraphics[width=\textwidth]{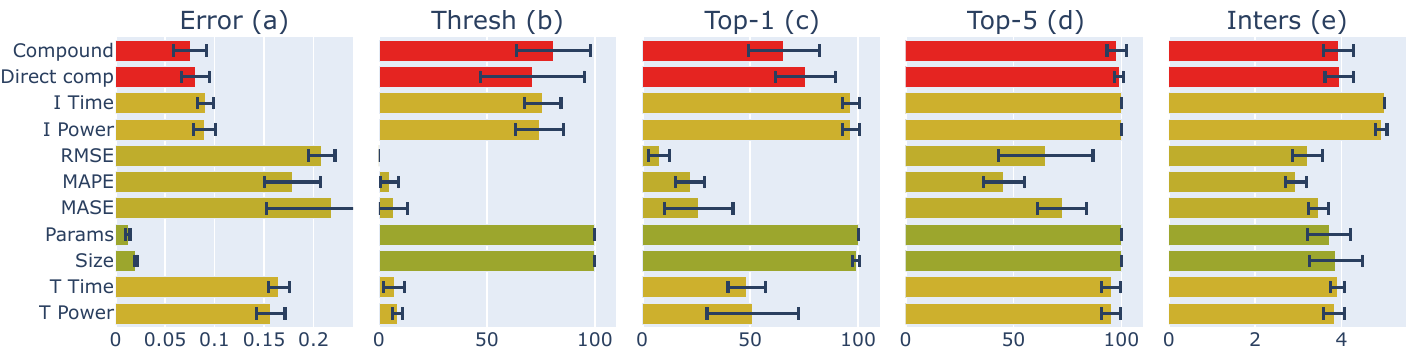}
    \caption{Quality of estimating properties and compound score. Some properties like errors are harder to predict, resulting in higher errors. The top-5 accuracy and intersection prove how testing the top-5 recommendations almost certainly retrieves the optimal solution. Colors indicate each property's weight in the compound score (red).}
    \label{fig:error_of_recommendation}
\end{figure*}

To demonstrate the explainability of \mname, Figure \ref{fig:explanations} shows by-product explanations of an exemplary recommendation.
It firstly informs on which properties are most supportive of the model recommendation, measured by their normalized importance for the compound score estimate.
For this example, we see that the estimated MASE makes FFO a favorable choice for the Car Parts data.
Our method also allows to dig deeper by querying the interpretable property estimation models for meta-feature importance.
In this case, we see that -- besides model choice -- the number of series and average minimum value mostly affected the MASE estimate.
Lastly, we discuss the quality of estimating properties and compound scores.
We assess the error (a, recall also Equation \ref{eq:estimation_error}), the accuracy of scoring errors below a threshold of 0.1 (b), top-1 (c) \& top-5 (d) accuracy of predicting the optimal model, and the intersection size of top-5 recommendation and top-5 true best models (e), as defined in Table \ref{tab:metalearn_errors}.
These error measures and standard deviations across all data sets are provided in Figure \ref{fig:error_of_recommendation}, with bar colors indicating the associated property impact on the compound score (colored red).
As expected, errors and resource demand for training are the hardest to predict, while model complexity seems to behave rather deterministic.
The top-5 accuracy and intersection results support our analysis of Figure \ref{fig:convergence} -- testing the five best recommendations has a high chance of finding the optimal model.
We also see empirical evidence that compositional PCR selection is superior to directly estimating the compound score, as questioned in Section \ref{ssec:meth:modelselection}.

\section{Conclusion}
To conclude, we introduced \mname, which -- to the best of our knowledge -- is the first automated, explainable and resource-aware take on model selection.
While we successfully used it to select DNNs for forecasting time series, our methodology can be easily generalized to other domains.
The meta-learning paradigm of our method pays close attention to different PCR objectives, which describe model behavior in terms of predictive error, complexity, and resources.
Calculating index values and compound scores makes these properties comparable -- here we improve existing work on assessing the efficiency of ML.
Model recommendations are accompanied by multi-level explanations, and the interactiveness of our framework allows users to control and understand the selection process.
Our experimental evaluation gives evidence of \mname's practical feasibility.
It outperforms competing approaches and achieves near-to-optimal predictive quality while only requiring a fraction of the computational effort.
We deem our approach highly beneficial for the domain of meta-learning and model selection, as well as forecasting.
For future work, we intend to apply \mname~in even more domains and environments.
With our work, we hope to contribute to making both ML and time series forecasting in particular, more resource-aware and trustworthy.

\subsection*{Acknowledgement}
This research has been funded by the Federal Ministry of Education and Research of Germany
and the state of North Rhine-Westphalia as part of the Lamarr Institute for
Machine Learning and Artificial Intelligence.

\bibliographystyle{splncs04}
\bibliography{references}

\end{document}